\documentclass[letterpaper]{article} 
\usepackage{aaai25}  
\usepackage{times}  
\usepackage{helvet}  
\usepackage{courier}  
\usepackage[hyphens]{url}  
\usepackage{graphicx} 
\usepackage{natbib}  
\usepackage{caption} 
\usepackage{arydshln}
\usepackage{algorithm}
\usepackage{algorithmic}
\usepackage{amsmath}
\usepackage{pdfpages}

\usepackage{newfloat}
\usepackage{listings}
\DeclareCaptionStyle{ruled}{labelfont=normalfont,labelsep=colon,strut=off} 
\floatstyle{ruled}
\newfloat{listing}{tb}{lst}{}
\floatname{listing}{Listing}
\title{Label-Efficient Data Augmentation with Video Diffusion Models for Guidewire Segmentation in Cardiac Fluoroscopy}
\author {
    Shaoyan Pan\textsuperscript{\rm 1}\thanks{The work conducted by the first author was carried out during an internship at United Imaging Intelligence.},
    Yikang Liu\textsuperscript{\rm 2},
    Lin Zhao\textsuperscript{\rm 2},
    Eric Z. Chen\index{Chen,Eric Z.}\textsuperscript{\rm 2},
    Xiao Chen\textsuperscript{\rm 2},
    Terrence Chen\textsuperscript{\rm 2},
    Shanhui Sun\textsuperscript{\rm 2}
}
\affiliations {
    \textsuperscript{\rm 1}Emory University, Atlanta, GA, USA\\
    \textsuperscript{\rm 2}United Imaging Intelligence, Boston, MA, USA\\
    pshaoya@emory.edu, \{yikang.liu, lin.zhao01, zhang.chen, xiao.chen01, terrence.chen, shanhui.sun\}@uii-ai.com
}

\usepackage{bibentry}

\begin{document}

\maketitle

\begin{abstract}
The accurate segmentation of guidewires in interventional cardiac fluoroscopy videos is crucial for computer-aided navigation tasks. Although deep learning methods have demonstrated high accuracy and robustness in wire segmentation, they require substantial annotated datasets for generalizability, underscoring the need for extensive labeled data to enhance model performance. To address this challenge, we propose the Segmentation-guided Frame-consistency Video Diffusion Model (SF-VD) to generate large collections of labeled fluoroscopy videos, augmenting the training data for wire segmentation networks. SF-VD leverages videos with limited annotations by independently modeling scene distribution and motion distribution. It first samples the scene distribution by generating 2D fluoroscopy images with wires positioned according to a specified input mask, and then samples the motion distribution by progressively generating subsequent frames, ensuring frame-to-frame coherence through a frame-consistency strategy. A segmentation-guided mechanism further refines the process by adjusting wire contrast, ensuring a diverse range of visibility in the synthesized image. Evaluation on a fluoroscopy dataset confirms the superior quality of the generated videos and shows significant improvements in guidewire segmentation.
\end{abstract}

%

\section{Introduction}
Guidewires are frequently used in interventional surgeries to navigate blood vessels and guide other medical devices, such as catheters, stents, or balloons, to precise locations within the body. Guidewire segmentation in X-ray fluroscopy videos is a critical task in computer-aided cardiac interventional procedures. It often serves as a prerequisite for functions such as 4D guidewire reconstruction \cite{recon1} and the co-registration of angiography and intravascular ultrasound images \cite{coreg}. This task is inherently challenging due to factors such as image noise, complex scenes, projective foreshortening, object occlusion, rapid guidewire motion, and the thin, elongated shape of the guidewire.

\begin{figure}[!h]
\centering
\includegraphics[width=\columnwidth]{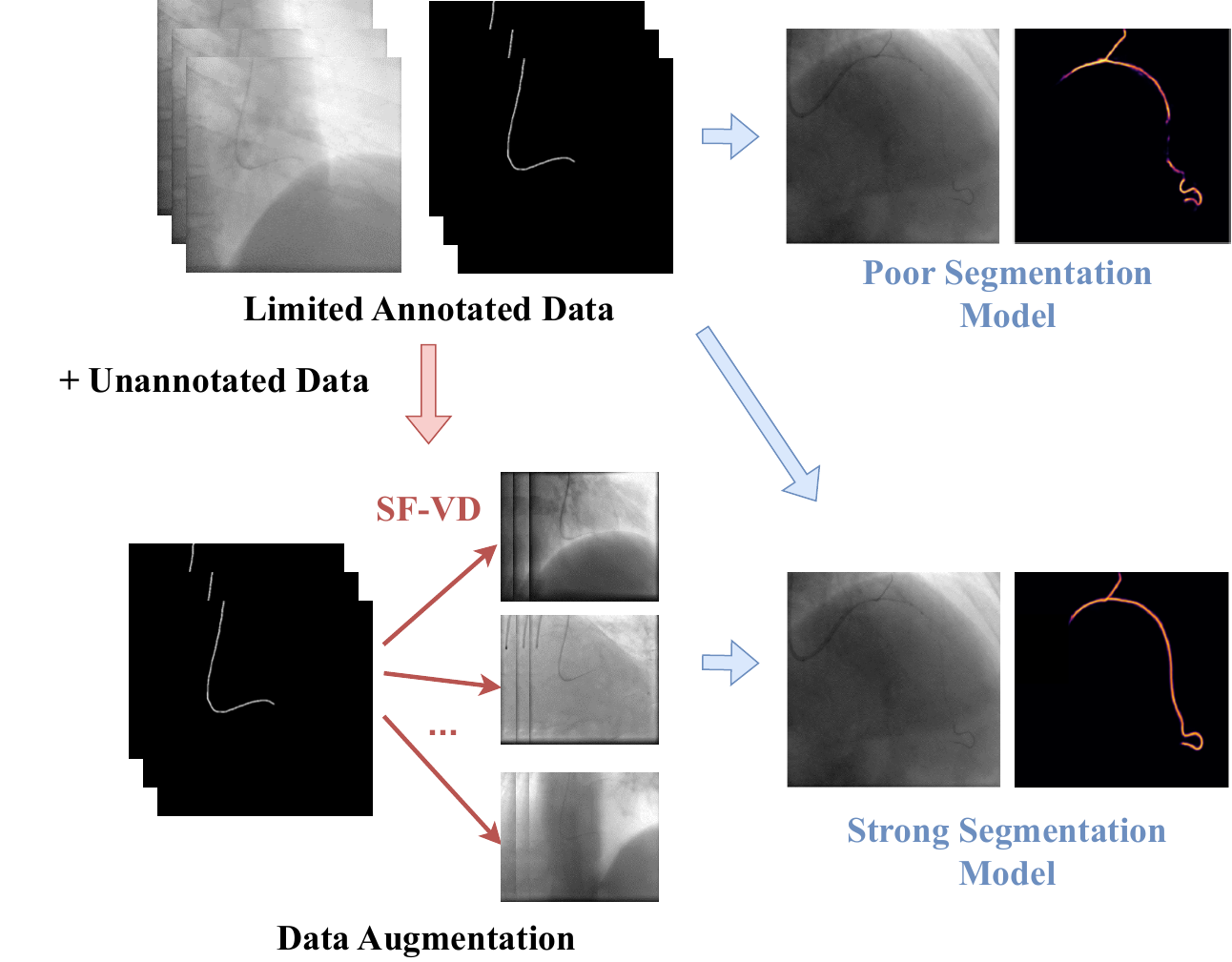}
\caption{SF-VD as a strong data augmentation tool for training guidewire segmention networks.} \label{teaser}
\end{figure}

Accurate and sufficient annotated data are essential for training guidewire segmentation models, but labeling thin and elongated guidewires is both error-prone and labor-intensive. To address this challenge, this paper proposes a method to generate synthetic videos paired with guidewire masks as augmented data, aimed at enhancing the performance of downstream segmentation models (Fig. \ref{teaser}). We trained a generative diffusion model (DM) on a small labeled dataset along with a larger unannotated dataset, enabling the generation of multiple annotated fluoroscopy videos with varying backgrounds and wire appearances from a given semantic wire mask. 

In medical data synthesis, Generative Adversarial Networks (GANs) \cite{ganGoodfellow} encounter challenges such as unstable training and limited video fidelity and diversity, particularly due to the complex distribution of medical images and small dataset sizes. Denoising Diffusion Probabilistic Models (DDPMs) \cite{ddpmho,ddpmsong} have shown promise in generating diverse medical images \cite{medicalddpmyu,medicalddpmdu,medicalddpmpan}. Similarly, video DDPMs \cite{videoddpmBlattmann,videoddpmBlattmann2,videoddpmho} demonstrate potential in synthesizing medical videos conditioned on semantic contents \cite{ddpmlug,conditionddpmwang,medicalddpmdu}. 

However, our task of synthesizing fluroscopy videos for data augmentation in wire segmentation poses unique challenges and specific requirements. Firstly, it is challenging to acquire fluoroscopy images/videos at the scale of natural ones, and annotating wires in numerous videos is almost impractical. Therefore, it is beneficial to minimize the labeling effort while preserving the fidelity and variability of the synthesized videos.  Secondly, since guidewires are extremely thin (typically 2-3 pixels wide) and tend to be occluded by other objects or noise, detection often relies on the contrast between fast-moving guidewires and the slower-moving background, which needs to be reflected in synthesized videos. Thirdly, the nuanced and varied appearance of wires against complex backgrounds in fluoroscopy videos requires a conditioning approach that accurately captures wire masks and generates diverse wire appearances. Inadequate representation of these masks compromises the fidelity and variability of wire depiction in the synthesized videos, reducing their value for segmentation tasks. 

We propose a label-efficient data augmentation method for guidewire segmentation in cardiac fluroscopy videos, named Segmentation-guided Frame-consistency Video Diffussion Model (SF-VD). First, we use two separate 2D diffusion models (DMs) to independently learn scene distribution and cross-frame motion: the first model synthesizes a static image conditioned on a wire mask, while the second model generates subsequent frames based on the static image and the wire masks for those frames. This strategy allows the first model to effectively leverage data with limited labeled frames, while the second model ensures cross-frame consistency by training on fully annotated videos. Together, these models generate more diverse backgrounds than would be possible using only fully annotated video datasets. Additionally, the second model employs a ``frame-consistency" sampling strategy that produces more realistic and diverse videos compared to 3D models, especially given our limited data. The second key idea is to modulate wire contrast and enhance the variability of wire appearance and visibility in synthetic frames. This is achieved by using a segmentation model to guide the diffusion reverse process, similar to the approach used in classifier guidance \cite{imageguidedddpm}. 

We benchmarked the proposed method against various data augmentation methods including other video diffusion models, using several segmentation model architectures, which included both generic state-of-the-art (SOTA) models and those specifically designed for guidewire segmentation. Our results demonstrate that our method outperforms other data augmentation approaches and enhances the segmentation performance of all tested models.

In summary, the key contributions of this work are:
\begin{itemize}
\item Introduced SF-VD, a label-efficient video synthesis model that enhances downstream wire segmentation performance. To the best of our knowledge, this is the first work on fluoroscopy video synthesis and the first to use a generative model for data augmentation in guidewire segmentation.
\item Proposed a two-model strategy to independently learn scene and motion distributions, addressing the challenges of limited medical data and guidewire annotations.
\item Proposed a frame-consistency strategy to ensure coherent motion and content across fluoroscopy frames.
\item Introduced diffusion guidance by a segmentation network to adjust guidewire contrast, improving visualization and diversity in fluoroscopy videos.
\end{itemize}

\begin{figure*}[!h]
\includegraphics[width=\textwidth]{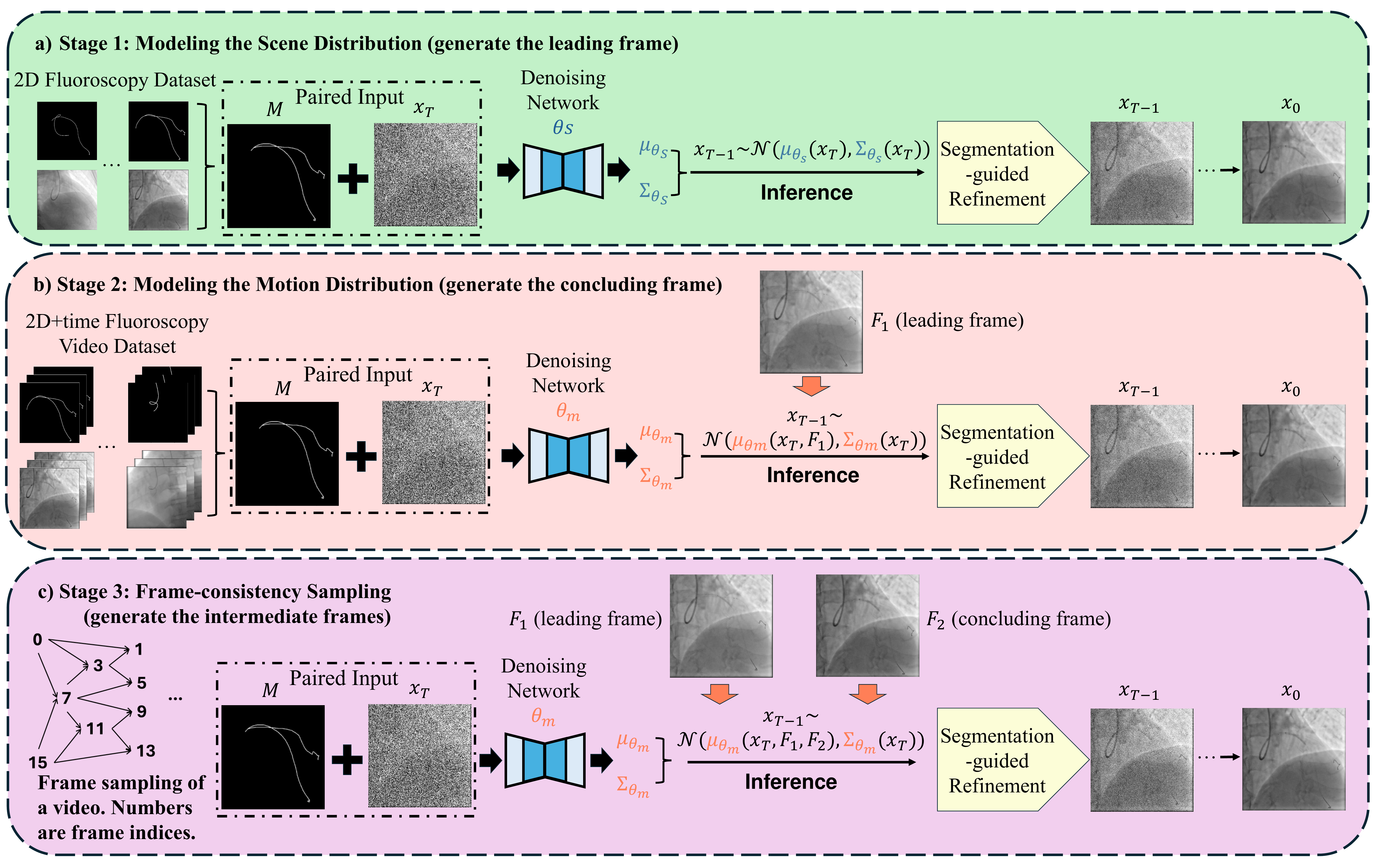}
\caption{Overview of the three-stage approach in SF-VD.} \label{method}
\end{figure*}
\section{Related Work}
\subsubsection{Video Diffusion Model}
Recent advancements in DMs have driven progress in video generation. Several studies \cite{videoddpmho,3DvideoDDPM3,liu2024sora} train 3D DMs to model video sequences as volumetric images. However, optimizing these models is challenging due to the high-dimensional nature of the data, which requires a large amount of video data and compute \cite{chen2024videocrafter2}. In response, some researchers used pretrained 2D image DMs for consistent frame generation. They either integrated additional structures to fine-tune the pretrained image DMs \cite{2DvideoDDPM1, 2DvideoDDPM3, 2DvideoDDPM8, 2DvideoDDPM7, videoddpmBlattmann2}, or employed a completely training-free approach \cite{2DvideoDDPM5, 2DvideoDDPM4}, using the image DMs as a generative prior to generate videos. These methods reduce optimization challenges and improve video fidelity. However, directly applying these techniques to our problem is not optimal, since fine-tuning pretrained image DMs still requires a substantial number of labeled videos, which are costly to obtain for medical data. Moreover, precise pixel-level conditioning is crucial for our task, and maintaining this accuracy during sampling from a pretrained image DM is not straightforward. To address these challenges, we propose training a DM from scratch with pixel-wise wire conditioning and a two-model strategy, which independently models scene and motion distributions to reduce the amount of required data.

\subsubsection{Conditional Diffusion Model}
Recent advancements in conditional DMs have significantly enhanced personalized, task-specific image and video generation by optimizing condition integration for better control. Key approaches include `network injection' DMs, which embed conditions within the denoising network for controlled generation \cite{conditionddpmwang,conditiondiffusioninnetwork3}, and `process injection' DDPMs, which modify the diffusion trajectory by incorporating conditions directly into the diffusion steps \cite{ddpmlug,conditiondiffusioninprocess2,conditiondiffusioninprocess3}. Classifier-guided-based DMs utilize an auxiliary network to preprocess conditions before injection, offering precise guidance \cite{imageguidedddpm,conditiondiffusionextranetwork1,controlnet}. Classifier-free DMs \cite{classifierfree,conditionaldiffusionclassifierfree1} employ a simpler architecture without explicit condition encoding, making them suitable for applications such as text-guided, segmentation-guided generation, and colorization. In the medical field, these methods have been adapted for conditional image generation, such as SegGuidedDiff \cite{conditiondiffusionsegdiff} for generating breast MRI and abdominal CT images with organ segmentation maps, Med-DDPM \cite{conditiondiffusionMEDDDPM} for generating 3D semantic brain MRI with brain organ segmentation maps, and ArSDM \cite{medicalddpmdu} for producing colonoscopy images with polyps using polyp segmentation maps. These advancements reveal the potential of conditional DMs to generate medical images with precise, pixel-wise correspondence to specified segmentation maps. Building on this, our work introduces SF-VD, a model that uses wire masks to generate videos with clear wires precisely aligned to the given masks. In addition, SF-VD incorporates a controllable conditioning strategy that modulates wire contrast, enhancing the variability of wire appearance and visibility across synthetic frames.

\subsubsection{Guidewire Segmentation in Fluoroscopy Videos}
Advancements in deep learning, particularly with convolutional neural networks (CNNs) \cite{segmentationvnet,segmentationfabian} and vision Transformers \cite{segmentationTang,segmentationPan}, have shown promise in accurately segmenting medical objects, including organs and interventional devices \cite{segmentationbaseline3,segmentationinstructment1}. Innovative solutions such as pyramid attention recurrent networks (PARN) \cite{segmentationpyramid_attention}, temporal Transformer network (TTN) \cite{segmentationtemporary}, and 4D recurrent networks (4D-RCN) \cite{segmentation4D} have been proposed to enhance feature capture across video frames, leading to improved guidewire segmentation. Additional strategies like real-time guidewire network (RTGN) using a preliminary guidewire detection network for bounding box setup \cite{segmentationrealtime}, and Ariadne+ \cite{segmentationariande}, which applies graph theory for post-process refinement and false negative reduction, have been proposed. These models focus on development of neural network architectures and segmentation algorithms, but the scarcity of annotated training videos is overlooked, which often limits their performance on out-of-distribution videos. We propose a label-efficient data augmentation method using conditional DMs to improve segmentation network training with synthetic data that cover a wide range of guidewire contrast, background appearances, and motions.

\section{Methodology}
\subsubsection{Overview}
The aim of the proposed video generation method, SF-VD, is to take a sequence of guidewire masks (as illustrated in Fig. \ref{teaser}) and synthesize fluoroscopy videos with guidewires at the specified pixels, which exhibit good variation, fidelity, and cross-frame consistency. The method aims to efficiently utilize fluoroscopy videos and mask annotations, given their limited availability, and to serve as a data augmentation tool for training wire segmentation models.

To address of the problem of overfitting when directly modeling the complex condition distribution of fluoroscopy videos, we propose to decompose the distribution $P(\mathbf{V}|\mathbf{M})$ ($\mathbf{V}=\{I^{(0)}...I^{(N-1)}\}$ is a $N$-frame video and $\mathbf{M}=\{M^{(0)}...M^{(N-1)}\}$ is its corresponding guidewire masks) into two components: a scene distribution $P(I^{(0)}|M^{(0)})$, which accounts for factors such as imaging angles, imaging parameters, objects in view, and patient anatomy; and a motion distribution $P(I^{(1)}...I^{(N-1)}|M^{(1)}...M^{(N-1)}, I^{(0)})$, which captures object motion due to heartbeats and breathing. The strategy also helps to leverage the video with partially labeled frames, which can be used to learn the scene distribution.

We trained two 2D DMs to independently model each component and employed a three-stage approach (one stage to sample the scene distribution; two stages to sample the motion distribution) to synthesize videos, as shown in Fig. \ref{method}. In addition, a segmentation-guided mechanism was introduced to refine the diffusion reverse process throughout, ensuring controllable wire depiction across frames.





\subsubsection{Modeling the Scene Distribution}
Modeling a distribution with DMs involve two primary Gaussian processes: forward and reverse diffusion\cite{ddpmnic}. In the forward process, an uncorrupted image \( x_0 \) undergoes \( T \) diffusion steps, progressively incorporating noise \(\epsilon\), resulting in a sequence of increasingly noisy frames: \(\{x_0, \ldots, x_T\}\). This process is modeled as a Gaussian process with a specified noise schedule \(\{\alpha_0, \ldots, \alpha_T\}\): $x_t \sim \mathcal{N}(\sqrt{\overline{\alpha}_t} x_0, (1 - \overline{\alpha}_t)\mathbf{I})$,
where \(\overline{\alpha}_t\) is the product of the noise coefficients up to time \(t\), \(\overline{\alpha}_t = \prod_{i=1}^{t} \alpha_i\).  

Next, a network parameterized by \(\theta\) is trained to reverse the forward process, starting from a noise sample and sequentially removing noise \(\epsilon\) to recover the original data. 

To model the scene distribution of fluoroscopy videos, we want to generate an image conditioned on a guidewire mask. Given a noisy sample \( x_t \) and the time step \( t \), the model predicts the previous less noisy sample \( x_{t-1} \) with the mask \( M \):
\begin{equation}
  \begin{aligned}
  x_{t-1} \sim p(x_t \mid M) &= \mathcal{N}(\mu_\theta(x_t, t, M), \sigma^2_\theta(x_t, t, M)\mathbf{I}) \\
  \text{where} \quad \mu_\theta(x_t, t, M) &= \frac{1}{\sqrt{\alpha_t}}\left(x_t - \frac{1-\alpha_t}{\sqrt{1-\overline{\alpha}_t}}\epsilon_{\theta}(x_t, t, M)\right) \\
  \text{and} \quad \sigma^2_\theta(x_t, t, M) &= \exp(v_\theta v_f +  (1 - v_\theta)v_r)
  \end{aligned}
  \label{eq:ddpm}
\end{equation}
Here, the condition mask \( M \) is channel-wise concatenated with \( x_t \), and fed to the network along with the time step \(t\) to estimate the noise \(\epsilon_{\theta}\) and variance \( v_\theta \). \( v_f \) and \( v_r \) are two variance-related constants defined in IDDPM \cite{ddpmnic}. The optimization for \(\mu_\theta\) and \(\sigma_\theta\) follows IDDPM, as shown in Appendix A.

The network was trained on a mixture of annotated and unannotated images to increase background variability. The strategy aligns with the formulation of the classifier-free (CF-) DDPM \cite{classifierfree}, as the network simultaneously models both the unconditional and conditional image distributions. CF-DDPM approximates the conditional score function as:
\begin{equation}
  \begin{aligned}
  \nabla_{x_t}& \log p(x_t \mid M) = \nabla_{x_t} \log p(x_t) + \nabla_{x_t} \log p(M \mid x_t) \\
  \approx & \nabla_{x_t} \log p(x_t ) + \omega (\nabla_{x_t} \log p(x_t \mid M) - \nabla_{x_t} \log p(x_t))
  \end{aligned}
  \label{eq:classifier-free-stage1}
\end{equation}
Unlike the original CF-DDPM where variance is constant, we incorporate a predicted variance to express the score function as:
\begin{equation}
  \nabla_{x_t} \log p(x_t \mid M) = \frac{\epsilon_{\theta}(x_t, t, M)}{\sigma_\theta(x_t, t, M)}
  \label{eq:score-epsilon-sigma}
\end{equation}

Rewriting Eqs. \ref{eq:classifier-free-stage1} and \ref{eq:score-epsilon-sigma} into the format of Eq. \ref{eq:ddpm}, we have the updated noise as:
\begin{equation}
  \bar{\epsilon}_{\theta}(x_t, t, M) = (1-\omega) \epsilon_{\theta}(x_t, t) + \omega \epsilon_{\theta}(x_t, t, M)
\end{equation}
, where \( \omega \) is empirically selected as 0.7. 
For the variance $\sigma^2_\theta(x_t, t)$, we retain the original form \( \sigma^2_\theta(x_t, t, M) \) since the updated form does not improve performance, likely because the unconditional variance is difficult to estimate accurately. 

\begin{figure*}[!ht]
\includegraphics[width=\textwidth]{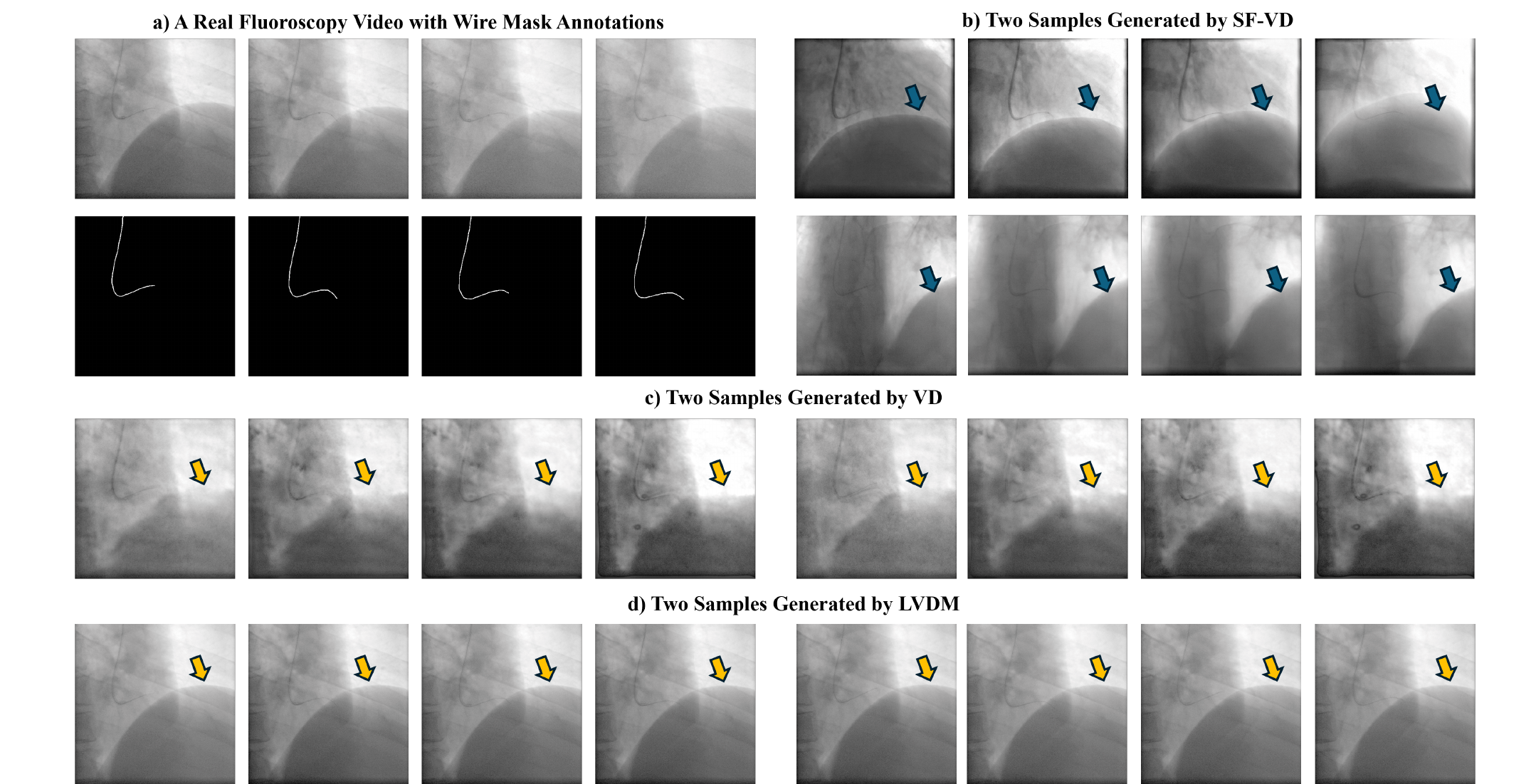}
\caption{Video Quality Comparison among SF-VD and Benchmarked Video Diffusion Models. Each sample includes a sequence of 4 frames, displayed from left to right, with the first frame on the left and the last frame on the right. The blue arrows indicate obvious diaphragm motions, and yellow arrows indicate less apparent motions} \label{diffusion-quality}
\end{figure*}
\begin{figure*}[!ht]
\includegraphics[width=\textwidth]{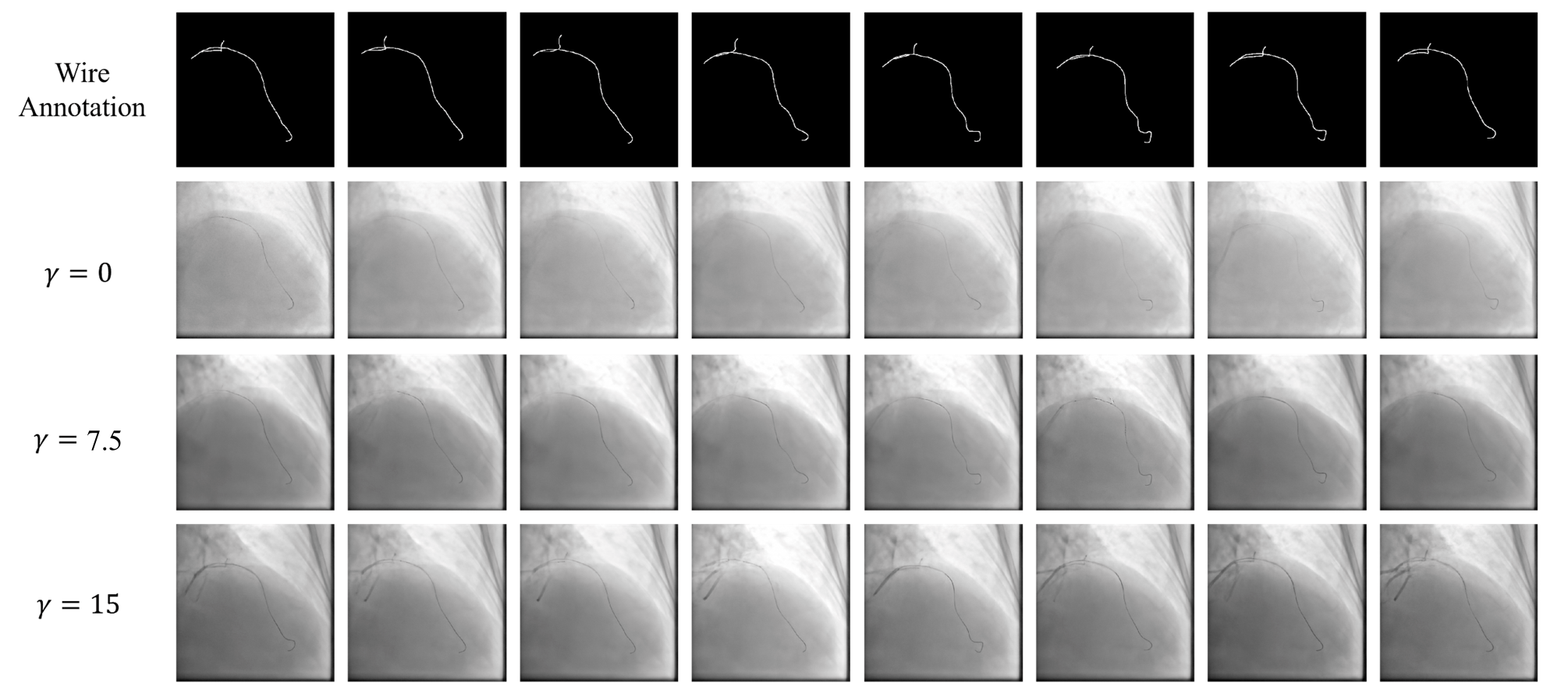}
\caption{The Segmentation-guided Mechanism Generates Videos with Wires with Various Contrasts. Each row shows a video sequence from the first frame on the left to the last frame on the right.} \label{fig-segmentation-guide}
\end{figure*}

\subsubsection{Modeling the Motion Distribution}
To model the motion distribution, we propose to generate another frame based on a previously generated frame and a mask with a 2D DM. Note that modeling motion in cardiac fluoroscopy videos is typically simpler than in natural videos because, in fluoroscopy, objects tend to stay within the field of view and their appearance remains relatively stable. Therefore, a 2D conditional DM is sufficient for the task and helps prevent overfitting with limited training data.

Similar to Eq. \ref{eq:classifier-free-stage1}, with the conditions of an initial frame \( F \) and a mask \( M \), we have:

\begin{equation}
  \begin{aligned}
  \nabla_{x_t} \log& p(x_t \mid F, M) \\
   = & \nabla_{x_t} \log p(x_t \mid M) + \nabla_{x_t} \log p(F \mid x_t, M) \\
  \approx & \nabla_{x_t} \log p(x_t \mid M) +  \omega (\nabla_{x_t} \log p(x_t \mid F, M) \\ 
  -& \nabla_{x_t} \log p(x_t \mid M))
  \end{aligned}
  \label{eq:classifier-free-stage2}
\end{equation}
Similarly, the predicted variance in Eq. \ref{eq:ddpm} is \( \sigma^2_\theta(x_t, t, F, M) \), and the predicted noise is:
\begin{equation}
  \bar{\epsilon}_{\theta}(x_t, t, M, F) = (1-\omega) \epsilon_{\theta}(x_t, t, M) + \omega \epsilon_{\theta}(x_t, t, M, F)
  \label{eq:epsilon-F}
\end{equation}

\subsubsection{Frame-consistency Sampling}
The networks described above enable us to generate the first frame of a video, followed by each subsequent frame sequentially. However, this approach produced videos with jittery background motion, negatively impacting segmentation performance (see Appendix D). To address this, we introduce ``frame-consistency" sampling, where a frame is generated with the conditions of the preceding and following frames. Denoting these frames as $F_1$ and $F_2$, similar to Eqs. \ref{eq:classifier-free-stage1} and \ref{eq:classifier-free-stage2}, the conditional score function is:
\begin{equation}
  \begin{aligned}
  \nabla_{x_t} \log& p(x_t \mid F_1, F_2, M) \\
  =& \nabla_{x_t} \log p(x_t, F_1, F_2, M) - \nabla_{x_t} \log p(F_1, F_2, M) \\
  =& \nabla_{x_t} \log p(x_t \mid M) + \nabla_{x_t} \log p(F_1 \mid x_t, M) \\
  +& \nabla_{x_t} \log p(F_2 \mid x_t, F_1, M)
  \end{aligned}
\end{equation}
Similarly, the predicted noise is
\begin{equation}
  \begin{aligned}
  \bar{\epsilon}_{\theta}(x_t,& t, M, F_1, F_2) = (1-2\omega)\epsilon_{\theta}(x_t, t, M) \\
  +& \omega \epsilon_{\theta}(x_t, t, M, F_1) + \omega \epsilon_{\theta}(x_t, t, M, F_2)
  \end{aligned}
  \label{eq:epsilon-2F}
\end{equation}

\subsubsection{A Three-stage Approach}
Taken together, as shown in Fig. \ref{method}, we first use the scene-distribution model $\theta_s$ to generate the first (leading) frame, followed by the motion-distribution model $\theta_m$ to generate the last (concluding) frame based on the leading frame. Next, we use the frame-consistency sampling to generate the middle frame, conditioned on the leading and concluding frames. The remaining frames are generated in the same manner (the sampling order of a 16-frame video is shown on the left in Fig. \ref{method}c), by iteratively generating frames equidistant from the previously generated ones until all frames are completed.

In practice, the motion-distribution model $\theta_m$ also takes the frame distance between the generated frame and the conditional frame as inputs. $\omega$ in Eqs. \ref{eq:epsilon-F} and \ref{eq:epsilon-2F} were set to $-2.5$ and $-1.5$ when generating the concluding frame and intermediate frames, respectively.

\subsubsection{Segmentation-guided Mechanism}
To modulate the wire contrast in fluoroscopy videos, a segmentation network \( \psi \), trained on the 2D fluoroscopy dataset, was used as a sampling guidance. The segmentation-guided mechanism is integrated into the estimated mean in the reverse process as follows:
\begin{equation}
  \begin{aligned}
  \mu_{t_{final}} =& \mu(x_t,t,M, F_1, F_2) \\
  +& \gamma\sigma_\theta(x_t,t,M)^2\log p_\psi(M|x_t) 
  \end{aligned}
\end{equation}
where \( \gamma \) controls the strength of the wire contrast. Typically, \( \gamma \) is randomly chosen from 0 (no segmentation-guided enhancement) to 15 to maintain visual realism.

\section{Experiments}
\subsubsection{Datasets}
Due to the absence of a publicly available cardiac fluoroscopy dataset, all data were collected from hospitals and manually annotated. Two datasets were used in this study: 1) a fully annotated fluoroscopy video (F-video) dataset consists of 400 2D+time fluoroscopy videos of 16 frames per video, from patients in cardiac interventional surgeries. Each video was normalized to an intensity range of -1 to 1. All frames are annotated with guidewire delineations by certified professionals. The frames were resized to \(512 \times 512\) pixels. 2) A partially annotated fluoroscopy image (P-image) dataset includes 409 2D+time fluoroscopy videos. Of these, 81 videos contain partial frames with guidewire annotations, and the rest videos have no annotations at all. Each video was normalized to an intensity range of -1 to 1. We collect 4000 annotated frames and 10000 unannotated frames, with all images resized to \(512 \times 512\) pixels, as the final dataset.

SF-VD was trained using both the P-image and F-video datasets, with both used for the scene-distribution model $\theta_s$, and the latter for the motion-distribution model $\theta_m$. Please refer to Appendix B for implementation details.

\subsubsection{Video Quality Analysis}
We evaluated the quality of synthesized videos using three metrics: Fréchet Video Distance (FVD) \cite{FVD}, Diversity Score (DS) \cite{DSOS}, and Overfitting Score (OS) \cite{DSOS}. FVD assesses the realism and diversity of the videos, DS measures the diversity among the videos, and OS evaluates whether the videos replicate training data or generate novel content.

\subsubsection{Wire Segmentation Evaluation}
In addition, we evaluated the impact of fluoroscopy videos synthesized by SF-VD on a downstream wire segmentation task. Segmentation models were trained using 80\% of the F-video dataset, augmented with 1200 synthesized videos featuring random wire contrasts generated by SF-VD. Performance was validated on another 10\% of the F-video, and tested on the rest 10\% of the F-video. 

Metrics to evaluate segmentation performance include Dice Score (Dice) for region-overlap accuracy, Hausdorff Distance (HD) for boundary-deviation accuracy, Ground Truth to Result Error (G2RE) and Result to Ground Truth Error (R2GE) \cite{segmentationeva1} for directional distance errors, and sensitivity and precision to address ambiguity in wire width and annotation variability. A more detailed explanation of these metrics is provided in Appendix B.

Various generic 3D segmentation networks, including V-Net \cite{segmentationvnet}, Dynvnet \cite{segmentationfabian}, 3D Swin-Unetr \cite{segmentationhata}, and 3D Token-based MLP-Mixer \cite{segmentationPan}, were used in the downstream segmentation task. Models specifically designed for wire segmentation, such as PARN \cite{segmentationpyramid_attention}, RTGN \cite{segmentationrealtime}, TTN \cite{segmentationtemporary}, and Ariadne+ \cite{segmentationariande}, were also included. 

The efficacy of SF-VD augmentation was compared with other augmentation strategies, 
including geometric data augmentations (GA), pre-training methods such as Contrastive Learning (CL) \cite{CL} and Masked Volume Modeling (MVI) \cite{MVI}, and synthesized videos from Video Diffusion Model (VD) \cite{videoddpmho} and Latent Video Diffusion Model (LVDM) \cite{videoddpmlatent}. 

Please refer to Appendix B for additional details on computing infrastructure, network implementation, and training.

\section{Results}
\begin{table}[b]
  \caption{Comparison of the Fluoroscopy Video Quality Metrics for Different Methods. \textbf{BOLD} indicates the best result.}
  \label{tab:fluoroscopy_comparison}
  \centering
  \begin{tabular}{lccc}
  \hline
  Method        & FVD ($\downarrow$)   & DS ($\downarrow$)   & OS ($\downarrow$)  \\
  \hline
  SF-VD       & \textbf{24.38}                & \textbf{0.76$\pm$0.17}       & \textbf{0.53$\pm$0.20}      \\
  VD            & 56.49                & 0.89$\pm$0.07       & 0.93$\pm$0.02      \\
  LVDM          & 27.55                & 0.91$\pm$0.05       & 0.92$\pm$0.04      \\
  \hline
  \end{tabular}
\end{table}

Fig. \ref{diffusion-quality} displays synthesized videos from the SF-VD, VD, and LVDM models. SF-VD generates realistic videos with diverse anatomical backgrounds and guidewire appearances. In contrast, VD, a 3D DDPM, results in noisy backgrounds and structural artifacts, highlighting the challenges of high-dimensional video generation without extensive datasets. LVDM-generated videos are visually realistic but exhibit limited variability, suggesting potential overfitting to a small annotated dataset. Conversely, SF-VD effectively produces varied videos by leveraging unannotated data. In addition, SF-VD generates realistic diaphragm motions, a capability that other methods fail to achieve. Fig. \ref{fig-segmentation-guide} shows synthetic videos with different wire contrasts, achieved through the segmentation-guided mechanism, which enhances the generalization capability and performance of the downstream segmentation network (Appendix D).

\begin{table}
    \caption{Comparison of data augmentation methods for wire segmentation using Dice ($\uparrow$) and HD ($\downarrow$): results for generic segmentation networks (upper table) and networks tailored for wire segmentation (lower table). \textbf{BOLD} indicates the best result. If SF-VD is the best, a paired t-test was conducted to compare it with the competing methods. *indicates p $<$0.05.}
    \centering
\small
\begin{tabular}{lccc}
\hline
\textbf{Method} & \textbf{Dynvnet} & \textbf{SwinUnetr} & \textbf{MLP-Mixer} \\
\hline
\textbf{No aug} & 0.52* / 28.38* & 0.50* / 15.32* & 0.52* / 11.85* \\
\textbf{+GA}    & 0.53* / 23.32* & 0.51* / 15.50* & 0.52* / 9.43*  \\
\textbf{+CL}    & 0.54* / 22.39* & 0.52* / 12.00*  & 0.54* / 10.50*  \\
\textbf{+MVI}   & 0.56* / 23.92* & 0.51* / 8.23*  & 0.53* / 10.38* \\
\textbf{+VD}    & 0.58 / 6.39*  & 0.58 / 7.51*  & 0.60 / 4.82*  \\
\textbf{+LVDM}  & 0.55* / 4.83*  & 0.58* / 5.40*  & 0.55* / 4.32   \\
\hline
\textbf{+SF-VD} & \textbf{0.61 / 3.93} & \textbf{0.61 / 4.23} & \textbf{0.61 / 4.22} \\
\hline
\hline
\textbf{Method} & \textbf{PARN} & \textbf{RTGN} & \textbf{TTN} \\
\hline
\textbf{No aug} & 0.52* / 68.37 & 0.49* / 62.11* & 0.52* / 44.17  \\
\textbf{+GA}    & 0.54* / 29.38 & 0.50* / 45.02* & 0.55* / 23.59  \\
\textbf{+CL}    & 0.52* / 26.79 & 0.53* / 29.23* & 0.50* / 16.40  \\
\textbf{+MVI}   & 0.53* / 20.31 & 0.53* / 35.90* & 0.48* / 8.39   \\
\textbf{+VD}    & 0.56* / 12.57 & 0.62* / 5.90*  & 0.52* / 6.55   \\
\textbf{+LVDM}  & 0.60* / \textbf{10.28} & 0.61* / 2.77 & 0.52* / \textbf{6.21} \\
\hline
\textbf{+SF-VD} & \textbf{0.62} / 10.54 & \textbf{0.66} / \textbf{2.65} & \textbf{0.58} / 6.92 \\
\hline
\end{tabular}   
\label{tab:performance_comparison}
\end{table}

For quantitative evaluation, Table \ref{tab:fluoroscopy_comparison} shows that SF-VD achieves the highest FVD, indicating superior realism. Although it performs comparably to LVDM, SF-VD notably enhances video diversity, benefiting downstream segmentation tasks. 

Table \ref{tab:performance_comparison} shows that SF-VD significantly improves wire segmentation performance of various generic segmentation networks. It surpasses unaugmented models with significant improvements in both region-overlap accuracy (Dice) and boundary-deviation accuracy (HD). Compared to other augmentation methods (CL, MVI, GA, VD, and LVDM), SF-VD achieves superior or comparable Dice and HD. In Appendix C, SF-VD also shows better or comparable sensitivity, precision, and G2RE. In addition, SF-VD shows effectiveness in networks specifically designed for wire segmentation, generally outperforming competing augmentations in Dice, HD and other metrics (see Appendix C). This underscores SF-VD's robust performance across various segmentation challenges.

\subsubsection{Ablation Study}
We investigate the impact of Frame-Consistency (FC) and Segmentation-Guided (SG) strategies on the wire segmentation performance of V-Net. FC is designed to generate more realistic anatomical backgrounds and motion within videos, whereas SG focuses on modeling the diverse appearances of wires. Experiments without FC generate frames in a chronological order.

Results in Appendix D reveal that incorporating FC significantly improves metrics that measure region-overlap accuracy, including Dice, sensitivity, and precision. In contrast, distance-measuring metrics such as HD, G2RE, and R2GE show only minor changes with the addition of FC. Note that these distance-based metrics are more sensitive to small, isolated false-positive regions located far from the wires. Therefore, the likely explanation is that FC, by generating videos with more temporally coherent and realistic background, enables segmentation networks to learn features that better distinguish foreground and background motions, and thus reduce the extent of false positives and negatives. However, FC does not fully eliminate isolated false-positive regions. This could be because other structures in cardiac fluoroscopy, such as ribs, likely exhibit motion patterns distinct from the background while also presenting sharp edges, which may be misclassified as wires. 

SG, by contrast, greatly reduces HD and G2RE, effectively minimizing distance-based errors. It also improves R2GE, especially when FC is applied, although its influence on region-overlap metrics remains minor. This suggests that while SG does not substantially reduce false-positive or false-negative areas, it effectively eliminates isolated false-positive regions. A possible explanation is that by diversifying wire contrast in training data, SG helps segmentation networks learn more robust features for distinguishing wires from wire-like structures, such as rib edges. This finding highlights the importance of incorporating high variability in wire appearance within training data. 

The simultaneous application of FC and SG yields the best results for Dice, HD, sensitivity, precision, and G2RE. This demonstrates that combining FC and SG strategies enhances both region-based and boundary-based metrics, achieving a well-rounded improvement.

\subsubsection{Limitation and Future Works}
The proposed SF-VD is relatively inefficient due to its dual DDPM architecture. It requires ~40 seconds to produce a 16-frame video on an A100 GPU, limiting both the number of videos and frames generated. Improving efficiency through a single model to handle all stages or by integrating high-efficiency DDPMs, such as the One-step Consistency Model \cite{song2023consistency} or DiffuseVAE \cite{pandey2022diffusevae}, is crucial for efficient video generation within the proposed three-stage framework.

Additionally, current methods rely on existing wire mask sequences, limiting variability in wire motion. Future research will explore synthesizing both wire motion and background dynamics in the videos. For instance, synthesizing diverse wire mask sequences to guide video generation will allow for greater variability. 

We also acknowledge that a limited number of training videos may affect the quality and variability of the generated outputs, potentially impacting downstream segmentation. Future work will analyze the relationship between training data size and segmentation performance to optimize data requirements.

Lastly, we aim to expand SF-VD to generate videos of other interventional devices, including catheters, stents, and catheter tips, to enhance segmentation and tracking for these devices.

\section{Conclusion}
This study presents the SF-VD framework, a pioneering effort in medical video synthesis, aimed at creating a diverse library of fluoroscopy videos from existing guidewire annotations and augmenting the training data for deep learning models for wire segmentation. Our evaluations demonstrate the efficacy of these synthesized videos in improving deep learning-based wire segmentation. Future work will explore the potential of SF-VD framework for a broader range of interventional device-related tasks in fluoroscopy videos.

\bibliography{aaai25}



\section*{Supplementary material}

# Appendix

## A  Practical Optimization of Diffusion Process

To estimate the $\mu_\theta(x_t, t, M)$ as shown in Section "Methodology: Modeling the Scene Distribution", we need to optimize $\epsilon_\theta(x_t, t, M)$:

$$\arg\min_\theta L_\mu^\theta = \arg\min_\theta \text{MAE}(\epsilon_\theta, \epsilon_\theta(x_t, t, M)) \tag{1}$$

To estimate the $\sigma_\theta(x_t, t, M)$ also shown in Section "Methodology: Modeling the Scene Distribution", we need to optimize $v_\theta(x_t, t, M)$:

$$\arg\min_\theta L_\sigma^\theta = \arg\min_\theta (D_{KL}\left(\mathcal{N}(\mu, \sigma^2) \mid \mathcal{N}(\mu, \sigma_\theta^2(v_\theta(x_t, t, M))))\right) \tag{2}$$

# B Implementation Details of Experiments

## B.1 Metrics

Due to the thin and narrow structure of wires, we used a diverse set of metrics for comprehensive evaluation. Dice Score (Dice) assesses overall overlap between the prediction and ground truth, while Hausdorff Distance (HD) captures boundary accuracy by measuring the largest deviations. Ground Truth to Result Error (G2RE) and Result to Ground Truth Error (R2GE) (Wu et al., 2008) measure directional errors: G2RE highlights missed regions, and R2GE identifies over-segmented areas. Sensitivity and precision at a 3-pixel error threshold address ambiguity in wire width and annotation variability, where small discrepancies (e.g., a wire being 3 pixels wide but annotated as 2 pixels) are considered acceptable.

## B.2 Computing Infrastructure

Our experiments were implemented using the PyTorch 2.0.1 framework with CUDA 11.7, running in Python 3.8.0 on Ubuntu 20.04.3 LTS. The computations were performed on a system equipped with a single NVIDIA A100 GPU with 40GB of memory, and Intel Xeon Gold 6226R Processor.

## B.3 Network Implementation and Training

SF-VD was trained using both the P-image and F-video datasets, with both used for the scene-distribution model $\theta_s$, and the latter for the motion-distribution model $\theta_m$. SF-VD was optimized with the AdamW (Loshchilov and Hutter, 2017) optimizer at a learning rate of $10^{-4}$. Training was conducted for 200 epochs for 2D fluoroscopy image synthesis and 500 epochs for 2D+time fluoroscopy video synthesis. Each frame of the synthesis videos are generated with 256 timesteps.

A 2D segmentation network based on DynUnet, an adaptation of nnUnet (Isensee et al., 2020), was trained for the segmentation-guided mechanism. Training utilized all annotated frames from the P-image and F-video dataset with Mixup (Zhang et al., 2017) augmentation. The AdamW (Loshchilov and Hutter, 2017) optimizer was used with a learning rate of $1 \times 10^{-5}$, a batch size of 3, and training lasted for 500 epochs, achieving full convergence.

Training for all segmentation models followed the strategies used for the segmentation-guided mechanism model, with variations in the number of epochs. Models were trained for 1000 epochs, consistent with the pre-defined epoch count for Dynvnet. For CL and MVI, pre-training was performed for 600 epochs. Due to the larger dataset (more than four times larger), training epochs for models with SF-VD, VD (Ho et al., 2022), and LVDM augmentations were reduced to 250 epochs to ensure fair comparison.

# C  Full Evaluation for Downstream Segmentation Task using Synthetic Video Augmentation

Table 1: Comparison of Data Augmentation Methods for Wire Segmentation with Generic Segmentation Networks. **BOLD** indicates the best result. If SF-VD is the best, a paired t-test was conducted to compare it with the competing methods. *indicates $p < 0.05$.

| Method | DS (↑) | Sensitivity (↑) | Precision (↑) | HD (↓) | G2RE (↓) | R2GE (↓) |
|---|---|---|---|---|---|---|
| | | | Vnet | | | |
| No aug | 0.47±0.10* | 0.62±0.11* | 0.72±0.09* | 46.42±5.91* | 33.57±4.32* | 3.49±1.39 |
| +GA | 0.48±0.10* | 0.68±0.11* | 0.73±0.07* | 33.38±5.85* | 22.53±4.21* | 3.45±1.35 |
| +CL | 0.49±0.09* | 0.66±0.10* | 0.74±0.08* | 41.38±5.72* | 30.58±4.15* | 3.30±1.02 |
| +MVI | 0.48±0.13* | 0.65±0.10* | 0.73±0.08* | 39.32±5.82* | 31.58±4.15* | **3.45±1.01** |
| +VD | 0.57±0.09* | 0.83±0.06 | 0.94±0.02* | 7.05±0.09* | 5.52±1.21* | 3.80±1.00 |
| +LVDM | 0.56±0.06* | 0.83±0.07 | 0.93±0.02* | 7.53±1.02* | 5.79±1.01* | 3.85±1.06 |
| +SF-VD | **0.59±0.10** | **0.84±0.06** | **0.95±0.02** | **4.83±0.78** | **4.39±1.18** | 3.87±1.04 |
| | | | Dynvnet | | | |
| No aug | 0.52±0.09* | 0.72±0.11* | 0.78±0.10* | 28.38±3.39* | 26.38±4.11* | 9.32±1.58* |
| +GA | 0.53±0.09* | 0.75±0.17* | 0.79±0.12* | 23.32±3.35* | 23.28±4.02* | 9.02±1.59* |
| +CL | 0.54±0.09* | 0.76±0.14* | 0.81±0.11* | 22.39±3.11* | 19.23±4.03* | 9.05±1.55* |
| +MVI | 0.56±0.09* | 0.75±0.10* | 0.85±0.08* | 23.92±2.80* | 21.33±3.54* | 19.29±1.56* |
| +VD | 0.58±0.07 | 0.86±0.07 | 0.88±0.03 | 6.39±0.81* | 5.84±1.31* | 5.93±0.93* |
| +LVDM | 0.55±0.01* | 0.82±0.06* | **0.90±0.06** | 4.83±0.79* | 3.82±1.09 | 3.94±1.04* |
| +SF-VD | **0.61±0.09** | **0.87±0.07** | **0.90±0.03** | **3.93±0.58** | **3.12±0.34** | **3.21±1.39** |
| | | | SwinUnetr | | | |
| No aug | 0.50±0.11* | 0.56±0.15* | 0.82±0.05* | 15.32±1.92* | 13.47±1.11* | 8.92±1.23* |
| +GA | 0.51±0.09* | 0.57±0.15* | 0.83±0.05* | 15.50±1.93* | 7.24±1.12* | 8.90±1.23 |
| +CL | 0.52±0.13* | 0.58±0.14* | 0.84±0.05* | 12.00±1.52 | 6.82±1.09* | 8.70±1.22* |
| +MVI | 0.51±0.13* | 0.57±0.15* | 0.83±0.06* | 8.23±1.85* | 7.32±1.10* | 8.85±1.22* |
| +VD | 0.58±0.06 | 0.81±0.05 | 0.85±0.04* | 7.51±0.72* | 2.53±0.37 | 6.13±0.94* |
| +LVDM | 0.58±0.08* | **0.84±0.10** | 0.83±0.03* | 5.40±0.71* | 3.48±0.36* | 5.05±0.93* |
| +SF-VD | **0.61±0.09** | 0.82±0.10 | **0.92±0.04** | **4.23±0.72** | **2.38±0.37** | **3.21±0.94** |
| | | | MLP-Mixer | | | |
| No aug | 0.52±0.12* | 0.67±0.10* | 0.75±0.08* | 11.85±2.84* | 10.33±1.74* | 8.48±1.20* |
| +GA | 0.52±0.11* | 0.71±0.10* | 0.73±0.08* | 9.43±2.34* | 9.31±1.34* | 7.64±1.19* |
| +CL | 0.54±0.12 | 0.69±0.09* | 0.77±0.08* | 10.50±2.75 | 9.39±1.68 | 8.37±1.34* |
| +MVI | 0.53±0.12* | 0.68±0.12* | 0.76±0.08* | 10.38±2.78* | 7.38±1.29* | 8.35±1.28* |
| +VD | 0.60±0.07 | 0.84±0.13 | 0.91±0.03 | 4.82±0.68* | 4.30±0.34 | **3.90±1.07** |
| +LVDM | 0.55±0.06* | 0.81±0.13* | **0.92±0.03** | 4.32±0.68 | **3.21±0.34** | 3.95±1.15 |
| +SF-VD | **0.61±0.08** | **0.85±0.13** | **0.92±0.03** | **4.22±0.68** | 3.24±0.35 | 4.00±1.05 |

Table 2: Comparison of Data Augmentation Methods for Wire Segmentation with Networks Tailored for Guidewire Segmentation. **BOLD** indicates the best result. If SF-VD is the best, a paired t-test was conducted to compare it with the competing methods. *indicates p <0.05.

| Method | Dice (↑) | Sensitivity (↑) | Precision (↑) | HD (↓) | R2GE (↓) | G2RE (↓) |
|---|---|---|---|---|---|---|
| PARN | | | | | | |
| No aug | 0.52±0.05 | 0.74±0.11 | 0.74±0.10 | 68.37±13.45 | 21.59±5.93 | 97.25±10.48* |
| +GA | 0.54±0.03 | 0.83±0.08 | 0.70±0.11 | 29.38±7.59 | 12.02±2.36 | 31.28±3.40* |
| +CL | 0.52±0.08 | 0.86±0.03 | 0.65±0.19 | 26.79±8.54 | 14.24±2.98 | 27.48±3.48* |
| +MVI | 0.53±0.07 | 0.84±0.05 | 0.78±0.04 | 20.31±4.50 | 19.68±4.56 | 27.06±5.97* |
| +VD | 0.56±0.10 | 0.82±0.03 | 0.94±0.02 | 12.57±3.78 | 4.51±0.47 | 14.68±3.06* |
| +LVDM | 0.60±0.10 | **0.88±0.03** | 0.84±0.05 | **10.28±1.20** | 5.95±0.27 | 13.02±2.19* |
| +SF-VD | **0.62±0.07*** | 0.87±0.03 | **0.95±0.01*** | 10.54±1.36 | **1.37±0.21*** | **10.93±1.57** |
| RTGN | | | | | | |
| No aug | 0.49±0.11 | 0.72±0.23 | 0.70±0.17 | 62.11±12.09 | 72.32±10.39 | 30.38±9.36 |
| +GA | 0.50±0.12 | 0.75±0.09 | **0.87±0.06** | 45.02±10.87 | 54.73±10.96 | 50.28±15.83 |
| +CL | 0.53±0.15 | 0.79±0.08 | 0.85±0.03 | 29.23±12.39 | 32.91±8.69 | 24.89±7.92 |
| +MVI | 0.53±0.17 | 0.82±0.04 | 0.75±0.06 | 35.90±9.72 | 25.81±7.49 | 41.75±8.69 |
| +VD | 0.62±0.14 | 0.90±0.03 | 0.79±0.07 | 5.90±1.68 | 3.59±0.16 | 2.11±1.34 |
| +LVDM | 0.61±0.15 | **0.95±0.04** | 0.87±0.09 | 2.77±1.00 | 2.57±0.18 | **1.89±0.52** |
| +SF-VD | **0.66±0.07*** | 0.93±0.03 | 0.86±0.08 | **2.65±1.21*** | **2.09±0.17*** | 2.74±0.91 |
| TTN | | | | | | |
| No aug | 0.52±0.07 | 0.63±0.14 | 0.94±0.03 | 44.17±10.35 | 24.50±8.49 | 67.49±13.30 |
| +GA | 0.55±0.12 | 0.62±0.13 | **0.95±0.03** | 23.59±7.80 | 15.22±3.79 | 31.05±9.60 |
| +CL | 0.50±0.13 | 0.64±0.12 | 0.69±0.23 | 16.40±5.54 | 17.59±2.96 | 17.08±4.11 |
| +MVI | 0.48±0.17 | 0.70±0.10 | 0.75±0.16 | 8.39±2.30 | 10.49±2.16 | 10.94±2.14 |
| +VD | 0.52±0.13 | **0.87±0.03** | 0.86±0.05 | 6.55±1.98 | 3.84±0.48 | 10.73±2.88 |
| +LVDM | 0.52±0.13 | 0.85±0.02 | 0.89±0.03 | **6.21±1.96** | **2.55±0.36** | 8.93±3.72 |
| +SF-VD | **0.58±0.08*** | 0.85±0.04 | 0.89±0.04 | 6.92±2.08 | 2.78±0.30 | **8.86±4.48*** |
| 4D-RCN | | | | | | |
| No aug | 0.52±0.07 | 0.63±0.14 | 0.74±0.03 | 16.92±2.08 | 2.78±0.30 | 22.86±4.48 |
| +GA | 0.54±0.07 | 0.63±0.17 | 0.75±0.03 | 16.53±4.20 | 3.30±0.58 | 30.34±6.68 |
| +CL | 0.50±0.13 | 0.69±0.13 | 0.79±0.23 | 14.89±3.96 | 5.23±1.95 | 14.69±3.05 |
| +MVI | 0.53±0.11 | 0.80±0.06 | 0.82±0.07 | 8.39±2.30 | 4.44±1.53 | 10.94±2.14 |
| +VD | 0.58±0.06 | **0.88±0.05** | 0.86±0.05 | **12.59±2.70** | 2.86±0.25 | **14.21±3.91** |
| +LVDM | 0.64±0.02 | **0.88±0.03** | 0.89±0.03 | 14.72±2.97 | 4.69±0.39 | 18.22±3.05 |
| +SF-VD | **0.66±0.04*** | **0.88±0.03** | **0.92±0.08*** | 14.00±3.49 | **2.17±0.21*** | 17.86±2.09 |
| Ariadne+ | | | | | | |
| No aug | 0.40±0.19 | 0.48±0.21 | 0.98±0.00 | 54.22±11.34 | 2.51±0.49 | 63.27±12.39 |
| +GA | 0.45±0.16 | 0.50±0.22 | 0.95±0.01 | 55.76±11.39 | 4.26±0.63 | 64.29±11.79 |
| +CL | 0.32±0.19 | 0.42±0.29 | 0.82±0.05 | 67.79±14.59 | 11.53±2.87 | 81.06±13.90 |
| +MVI | 0.43±0.16 | 0.42±0.37 | 0.84±0.16 | 55.56±12.48 | 13.28±2.37 | 55.97±8.98 |
| +VD | 0.57±0.13 | 0.84±0.03 | 0.83±0.05 | 2.55±1.57 | 1.83±0.37 | 5.46±1.37 |
| +LVDM | 0.55±0.08 | **0.88±0.03** | 0.82±0.06 | 2.91±0.68 | 1.59±0.45 | 5.38±1.32 |
| +SF-VD | **0.61±0.10*** | **0.88±0.04** | **0.90±0.04*** | 2.28±0.36 | **1.32±0.54*** | **2.28±0.28*** |

# D  Ablation Studies for Frame-consistency and Segmentation-guided Strategies

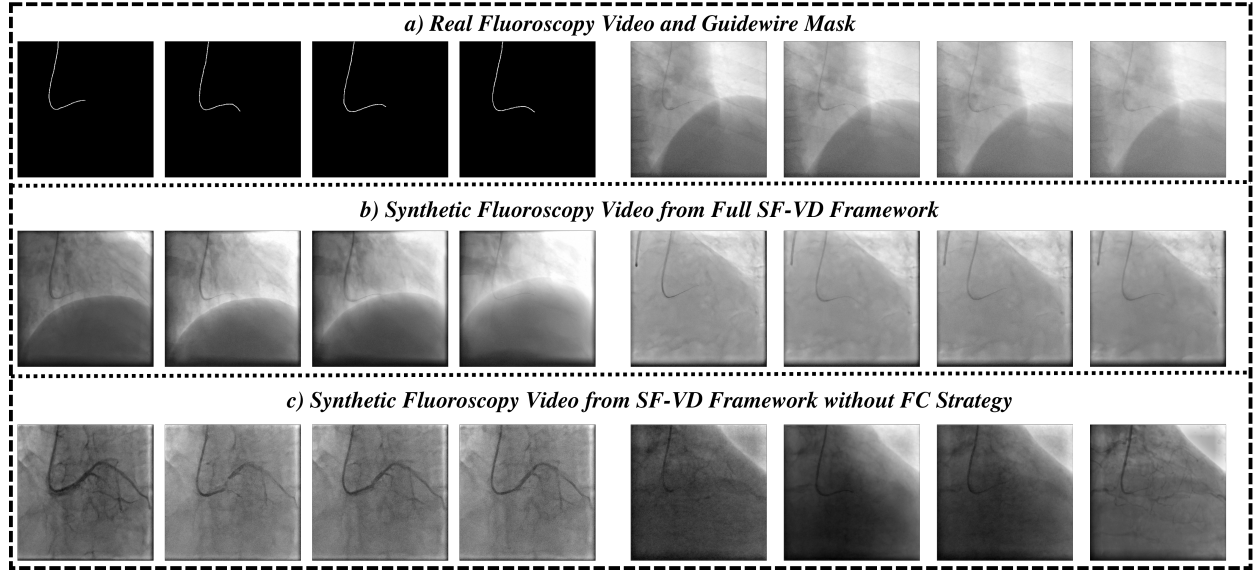

Figure 1: Visualization of Fluoroscopy Videos Generated by the Full SF-VD Framework and SF-VD without Frame-Consistency Strategy. The videos generated without the frame-consistency strategy exhibit inconsistencies across frames, particularly in anatomical structures, intensity, and motion, when compared to those generated by the full SF-VD framework.

Table 3: Quantitative evaluation of Frame-consistency (FC) and Segmentation-guided (SG) strategies on the SF-VD model's wire segmentation performance

| FC | SG | DS | Sensitivity | Precision | HD | G2RE | R2GE |
|---|---|---|---|---|---|---|---|
| No | No | $0.55 \pm 0.09$ | $0.76 \pm 0.09$ | $0.85 \pm 0.10$ | $8.78 \pm 1.84$ | $8.36 \pm 1.75$ | $3.53 \pm 1.50$ |
| No | Yes | $0.55 \pm 0.07$ | $0.73 \pm 0.11$ | $0.87 \pm 0.07$ | $6.11 \pm 0.74$ | $6.03 \pm 1.28$ | $3.23 \pm 1.10$ |
| Yes | No | $0.60 \pm 0.10$ | $0.83 \pm 0.03$ | $0.94 \pm 0.04$ | $8.92 \pm 2.03$ | $8.02 \pm 1.34$ | $5.43 \pm 1.32$ |
| Yes | Yes | $0.59 \pm 0.10$ | $0.84 \pm 0.06$ | $0.95 \pm 0.02$ | $4.83 \pm 0.78$ | $4.39 \pm 1.18$ | $3.87 \pm 1.04$ |

# E   Pseudocodes

**Algorithm 1:** SF-VD training

**Input:** #iteration for training: $n_1, n_2, n_3$, optimizer: $\Omega$, synthesis network: $\theta_s$ and $\theta_m$, Mask: $M$, 2D fluoroscopy image: $X$, leading frame: $F_1$, concluding frame: $F_2$, all zeros placeholder: $\emptyset$

1 **Stage 1: Leading Frame Synthesis**
2 **for** $i = 1, \ldots, n_1$ **do**
3      $M, X \sim$ Image-dataset$((M, X))$;
4      $t \sim$ Uniform$(\{1, \ldots, T\})$;
5      $\epsilon_t \sim \mathcal{N}(0, 1)$;
6      $X_t \leftarrow \sqrt{\bar{\alpha}_t} X + \sqrt{1 - \bar{\alpha}_t} \epsilon_t$;
7      50% chance:
8          $\epsilon_\theta, v_\theta = \theta_s(X_t, t)$ ($M$ is concatenated with $X_t$ and three $\emptyset$);
9      Another 50% chance:
10         $\epsilon_\theta, v_\theta = \theta_s(X_t, t, M)$ ($M$ is concatenated with $X_t$ and two $\emptyset$);
11      Use $\Omega$ to optimize $\theta_s$ based on argmin $L_\mu^\theta + L_\sigma^\theta$ using $\epsilon_\theta$ and $v_\theta$;

12 **Stage 2: Concluding Frame Synthesis**
13 **for** $i = 1, \ldots, n_2$ **do**
14      $M, F_1, X \sim$ Video-dataset$((M, F_1, X))$;
15      $t \sim$ Uniform$(\{1, \ldots, T\})$;
16      $\epsilon_t \sim \mathcal{N}(0, 1)$;
17      $X_t \leftarrow \sqrt{\bar{\alpha}_t} X + \sqrt{1 - \bar{\alpha}_t} \epsilon_t$;
18      50% chance:
19         $\epsilon_\theta, v_\theta = \theta_m(X_t, t, M)$ ($M$ is concatenated with $X_t$ and two $\emptyset$);
20      Another 50% chance:
21         $\epsilon_\theta, v_\theta = \theta_m(X_t, t, M, F_1)$ ($M$, $F_1$ concatenated with $X_t$ and one $\emptyset$);
22      Use $\Omega$ to optimize $\theta_m$ based on argmin $L_\mu^\theta + L_\sigma^\theta$ using $\epsilon_\theta$ and $v_\theta$;

23 **Stage 3: Intermediate Frame Synthesis**
24 **for** $i = 1, \ldots, n_3$ **do**
25      $M, F_1, F_2, X \sim$ Video-dataset$((M, F_1, F_2, X))$;
26      $t \sim$ Uniform$(\{1, \ldots, T\})$;
27      $\epsilon_t \sim \mathcal{N}(0, 1)$;
28      $X_t \leftarrow \sqrt{\bar{\alpha}_t} X + \sqrt{1 - \bar{\alpha}_t} \epsilon_t$;
29      33.3% chance:
30         $\epsilon_\theta, v_\theta = \theta_m(X_t, t, M)$ ($M$ is concatenated with $X_t$ and two $\emptyset$);
31      Another 33.3% chance:
32         $\epsilon_\theta, v_\theta = \theta_m(X_t, t, M, F_1)$ ($M$, $F_1$ is concatenated with $X_t$ and one $\emptyset$);
33      Final 33.3% chance:
34         $\epsilon_\theta, v_\theta = \theta_m(X_t, t, M, F_1, F_2)$ ($M$, $F_1$, $F_2$ is concatenated with $X_t$);
35      Use $\Omega$ to optimize $\theta_m$ based on argmin $L_\mu^\theta + L_\sigma^\theta$ using $\epsilon_\theta$ and $v_\theta$;

**Algorithm 2:** Segmentation Guide: Train a segmentation model

**Input:** #iteration for training: $n_4$, optimizer: $\Omega$, segmentation network: $\phi$, Mask: $M$, 2D fluoroscopy image or frame: $X$

1 **Stage 1: Leading Frame Synthesis**
2 **for** $i = 1, \ldots, n_4$ **do**
3      $M, X \sim$ [Image-dataset, Video-dataset]$([M, X])$;
4      $M_{\text{pred}} = \phi(X)$;
5      Use $\Omega$ to optimize $\phi$ based on Dice_loss$(M, M_{\text{pred}})$;

**Algorithm 3:** SF-VD inference

**Input:** Timesteps: $S$, score function frame weight: $\omega$, intermediate frame order: $\mathbf{l}$ (e.g., $\mathbf{l} = [15, 7, 3, 11, 1, 5, 9, 13, 2, 4, 6, 8, 10, 12]$)

1 **Stage 1: Leading Frame Synthesis**
2 $X_S \sim \mathcal{N}(0, I)$;
3 **for** $s = S, \ldots, n$ **do**
4     $\epsilon_s \sim \mathcal{N}(0, I)$;
5     $\epsilon_\theta^{\text{mask}} = \theta_S(X_S, s, M)$;
6     $\epsilon_\theta^{\text{no-mask}} = \theta_S(X_S, s)$;
7     $\sigma_\theta^2 = \theta_S(X_S, s, M)$;
8     $\bar{\epsilon}_\theta = (1 - \omega)\epsilon_\theta^{\text{mask}} + \omega\epsilon_\theta^{\text{no-mask}}$;
9     $X_{s-1} = \bar{\epsilon}_\theta + \sigma_\theta \epsilon_s$;
10 **return** $F_1 = X_0$
11 **Stage 2: Concluding Frame Synthesis**
12 $X_S \sim \mathcal{N}(0, I)$;
13 **for** $s = S, \ldots, n$ **do**
14     $\epsilon_s \sim \mathcal{N}(0, I)$;
15     $\epsilon_\theta^{\text{frame}} = \theta_m(X_S, s, M, F_1)$;
16     $\epsilon_\theta^{\text{no-frame}} = \theta_m(X_S, s, M)$;
17     $\sigma_\theta^2 = \theta_m(X_S, s, M)$;
18     $\bar{\epsilon}_\theta = (1 - \omega)\epsilon_\theta^{\text{frame}} + \omega\epsilon_\theta^{\text{no-frame}}$;
19     $X_{s-1} = \bar{\epsilon}_\theta + \sigma_\theta \epsilon_s$;
20 **return** $F_2 = X_0$
21 **Stage 3: Intermediate Frame Synthesis**
22 **foreach** $i \in \mathbf{l}$ **do**
23     $X_S \sim \mathcal{N}(0, I)$;
24     **for** $s = S, \ldots, n$ **do**
25         $\epsilon_s \sim \mathcal{N}(0, I)$;
26         $\epsilon_\theta^{\text{leading frame}} = \theta_m(X_S, s, M, F_1)$;
27         $\epsilon_\theta^{\text{concluding frame}} = \theta_m(X_S, s, M, F_2)$;
28         $\epsilon_\theta^{\text{no-frame}} = \theta_m(X_S, s, M, F_1)$;
29         $\sigma_\theta^2 = \theta_m(X_S, s, M) \cdot \text{Gradient}_\varphi(M \to X_S)$;
30         $\bar{\epsilon}_\theta = (1 - 2\omega)\epsilon_\theta^{\text{no-frame}} + \omega\epsilon_\theta^{\text{leading frame}} + \omega\epsilon_\theta^{\text{concluding frame}}$;
31         $X_{s-1} = \bar{\epsilon}_\theta + \sigma_\theta \epsilon_s$;
32     **return** $X_0$ *as intermediate frame* $F_l$, *set* $X_0$ *with* $F_2$ *or* $F_1$ *(depends on proximity for the next frame)*;
33 **return** *all frames* $F_1$ *(first frame)*, $F_2$ *(end frame), and* $F_l$ *to construct a video*.


\end{document}